\documentclass[times, review, 10pt]{elsarticle}
\usepackage{multirow}
\usepackage{threeparttable}
\usepackage{hyperref}
\usepackage[utf8]{inputenc}
\usepackage{bm}
\usepackage{amsmath}

\usepackage{algpseudocode}
\usepackage{algorithm}
\usepackage{mathtools}
\usepackage{amsmath}
\usepackage{amssymb}
\usepackage{multirow}
\usepackage{array,multirow}
\usepackage{xcolor}
\usepackage{rotating}

\usepackage{amsthm}
\usepackage{booktabs}
\usepackage{subcaption}
\usepackage{comment}
\usepackage{enumitem}
  \newtheoremstyle{dotless}{}{}{\itshape}{}{\bfseries}{}{ }{}
  \theoremstyle{dotless}

\usepackage{algpseudocode}
\usepackage{rotating}
\usepackage{caption}
\usepackage{algorithm}
\usepackage{mathtools}
\usepackage{stmaryrd}
\usepackage{multirow}
\usepackage{threeparttable}
\usepackage{xcolor}
\usepackage{booktabs}
\usepackage{hyperref}

\journal{}

\begin{document}

\begin{frontmatter}

\title{Locally Adaptive One-Class Classifier Fusion with Dynamic $\ell$p-Norm Constraints for Robust Anomaly Detection}
\author[bilkentcs]{Sepehr Nourmohammadi}
\author[bilkentcs]{Arda Sarp Yenicesu}
\author[bilkentcs]{Shervin Rahimzadeh Arashloo}
\author[bilkentcs]{Ozgur S. Oguz}
\address[bilkentcs]{Department of Computer Engineering, Bilkent University, Ankara, Turkey}
\cortext[mycorrespondingauthor]
{Corresponding author}\ead{ozgur.oguz@bilkent.edu.tr}

\begin{abstract}
This paper presents a novel approach to one-class classifier fusion through locally adaptive learning with dynamic $\ell$p-norm constraints. We introduce a framework that dynamically adjusts fusion weights based on local data characteristics, addressing fundamental challenges in ensemble-based anomaly detection. Our method incorporates an interior-point optimization technique that significantly improves computational efficiency compared to traditional Frank-Wolfe approaches, achieving up to 19-fold speed improvements in complex scenarios. The framework is extensively evaluated on standard UCI benchmark datasets and specialized temporal sequence datasets, demonstrating superior performance across diverse anomaly types. Statistical validation through Skillings-Mack tests confirms our method's significant advantages over existing approaches, with consistent top rankings in both pure and non-pure learning scenarios. The framework's ability to adapt to local data patterns while maintaining computational efficiency makes it particularly valuable for real-time applications where rapid and accurate anomaly detection is crucial.
\end{abstract}
\end{frontmatter}
%\linenumbers
\section{Introduction}
The integration of multiple classifiers through ensemble learning represents a fundamental advancement in machine learning, significantly enhancing classification performance \cite{kittler1998combining}. By incorporating diverse base learning algorithms, including decision trees, neural networks, and linear regression models, ensemble methods establish a comprehensive learning framework. These individual learners, each trained independently on labeled datasets, contribute distinct perspectives that collectively improve the overall predictive capabilities \cite{sagi2018ensemble}. 

The advantages of ensemble methodologies extend beyond basic accuracy improvements, offering particular strength in scenarios involving sparse or noise-contaminated data, reducing the probability of individual models becoming trapped in local optima, and facilitating more extensive exploration of potential solutions. Furthermore, these approaches demonstrate remarkable effectiveness in handling class imbalance issues \cite{japkowicz2002class}, show resilience to temporal variations in features and labels \cite{sagi2018ensemble}, and help prevent model overfitting.

Machine learning ensembles can be implemented through two distinct approaches: non-pure methods for multi-class classification and pure methods for one-class classification. Multi-class ensemble techniques improve model accuracy by combining multiple learning algorithms. One prominent approach is Bagging, exemplified by Random Forest algorithms \cite{breiman2001random}, which creates multiple decision trees trained on different data subsets and aggregates their predictions. This data diversity leads to more reliable and independent predictions. Alternatively, Boosting methodologies train sequential models that progressively address the shortcomings of previous iterations by prioritizing incorrectly classified samples. A classic implementation of this approach is AdaBoost \cite{freund1997decision}. Another ensemble strategy, Stacking, employs a hierarchical approach where various base models are trained and their predictions are synthesized using a higher-level model. This is demonstrated in Stacked Generalization \cite{wolpert1992stacked}, where a meta-model processes the outputs of base learners to generate final predictions. The voting classifier represents yet another ensemble method, where multiple algorithms' predictions are combined through majority voting to determine the final classification. This versatile approach can incorporate various machine learning algorithms to enhance overall predictive accuracy \cite{kuncheva2014combining}.

Ensemble learning demonstrates exceptional utility in One-Class Classification (OCC) applications. The primary objective of OCC is pattern recognition for identifying exceptional cases within datasets, a crucial capability across diverse domains. Due to the inherent scarcity of anomalous instances, developing comprehensive representations of anomaly classes presents significant challenges, making OCC methodologies particularly well-suited for anomaly detection applications \cite{krawczyk2018dynamic}. The implementation of ensemble techniques in OCC, including bagging, boosting, and stacking approaches, significantly improves the effectiveness of outlier identification processes.

Within the OCC framework, Bagging aims to have multiple OCC models trained on distinct subsamples of the training data, with their outputs subsequently combined. Boosting in the OCC context employs a weighted learning strategy, prioritizing samples that present greater classification difficulty. A notable implementation of this approach is demonstrated in Tax et al.'s Boosted SVM methodology for One-Class Classification \cite{tax2001uniform}. Another significant ensemble technique, stacking, integrates predictions from various classification or regression models through a higher-level learning algorithm. In this approach, base models are trained using the complete dataset, and their collective outputs inform the training of a meta-model. The effectiveness of stacking in OCC is well-documented, with Noto et al. \cite{noto2010anomaly} proposing an innovative two-tier ensemble framework. Their method first generates a diverse collection of one-class classifiers, followed by employing a binary classification algorithm to determine the optimal combination strategy for these base classifiers, effectively implementing stacking principles within OCC applications.

The inherent scarcity of anomalous instances often makes OCC approaches more suitable than traditional multi-class classification methods for anomaly detection. The limited availability of anomalous examples typically prevents the development of a comprehensive representation of the anomaly class, creating significant challenges for conventional binary classification methods. This limitation has led to an alternative strategy: establishing a decision boundary around normal instances, whereby any data points falling outside this defined boundary are categorized as anomalies \cite{seliya2021literature}.

Although not as widely implemented as their multi-class counterparts, ensemble techniques applied to One-Class Classification demonstrate significant potential in improving anomaly detection capabilities, especially in scenarios with limited anomalous examples. The integration of multiple models through ensemble approaches provides enhanced resilience against excessive adaptation to the normal class patterns, thereby improving the overall detection reliability.

Ensemble learning strategies can be categorized into two primary fusion paradigms: fixed-rule approaches and data-driven learning-based strategies. Fixed-rule fusion mechanisms, while prevalent in multi-classifier systems due to their computational efficiency and interpretability, present inherent limitations in their application scope. These approaches, such as the widely implemented sum rule, maximum rule, or majority voting, operate under predetermined combination protocols that treat all constituent models with uniform importance—an assumption that may not align with practical scenarios where certain classifiers demonstrate superior performance in specific  \cite{kittler1998combining_the}. The selection of appropriate fixed fusion strategies typically necessitates substantial domain expertise and often lacks the flexibility to adapt to varying data distributions or evolving pattern characteristics. 

Conversely, data-driven fusion strategies employ learning mechanisms to dynamically determine optimal combination weights or parameters based on the underlying data characteristics. These adaptive approaches can automatically adjust their fusion protocols through techniques such as weighted averaging, meta-learning, or stacked generalization, potentially offering more robust performance across diverse operational conditions. However, these learning-based methods generally require additional computational resources and training data to effectively learn the optimal fusion parameters \cite{naimi2018stacked}.

The challenges inherent in data-driven fusion strategies—particularly their computational requirements and need for training data—become especially pronounced when dealing with sequential data sequences. However, their ability to adaptively learn combination weights proves invaluable in temporal analysis scenarios, where the relationship between consecutive observations often defies fixed-rule approaches. This significance becomes apparent particularly in video analysis, introduces significant complexities in anomaly detection frameworks due to the intricate temporal dependencies between sequential frames. The fundamental challenge lies in scenarios where individual frames, while appearing nominal in isolated analysis, may constitute anomalous behavior when considered within their temporal context. This phenomenon is particularly evident in diverse application domains. 

In robotic systems monitoring, while instantaneous configurations may satisfy normal operational parameters, the temporal evolution of these states could represent suboptimal or potentially hazardous behavioral patterns that are only identifiable through local window assessment techniques. The necessity for temporal context integration is further exemplified in industrial automation, where robotic trajectories may momentarily intersect with valid state spaces while following globally anomalous paths. These scenarios emphasize the crucial requirement for incorporating temporal context in anomaly detection methodologies, as frame-independent analysis protocols may fail to capture subtle yet critical anomalies that manifest across temporal sequences. This necessitates the development of sophisticated detection frameworks that can effectively model and analyze both spatial and temporal relationships in sequential data.

Building upon our previous work \cite{nourmohammadi2025lp}, we propose an enhanced methodology for integrating multiple one-class classifiers through data-driven weighting mechanisms. This approach addresses fundamental challenges in OCC, particularly outlier sensitivity and skewed score distributions, while introducing local adaptivity—a crucial feature for complex scenarios requiring comprehensive local assessment, such as video frame anomaly detection. The methodology employs a hinge loss function within an $\ell_p$ norm constraint to optimize the ensemble weights, representing a significant advancement in OCC ensemble learning. In this work, we enhance our previous approach by implementing an interior point method, substantially improving computational efficiency. Furthermore, we introduce the novel public robotics anomaly dataset, LiRAnomaly, that encompasses both static and temporal anomalies. This dataset serves as a challenging benchmark for anomaly detection systems, particularly emphasizing the importance of locality-aware approaches in identifying complex behavioral patterns.

\subsection{Contributions}
The primary contributions of this research are:
\begin{itemize}
\item Development of a locally adaptive framework for ensemble integration, specifically designed to capture complex patterns prevalent in temporal sequences.
\item Implementation of an $\ell_p$ norm constraint methodology enhanced by an interior point optimization approach within conditional gradient descent, resulting in improved computational efficiency.
\item Introduction of a comprehensive public robotics anomaly dataset that encompasses both static and temporal anomalies, providing a robust benchmark for evaluating anomaly detection systems.
\end{itemize}
\subsection{Organization}
The remainder of this paper is organized as follows: Section 2 presents a comprehensive review of classifier fusion methodologies in both multi-class and one-class classification domains, along with an examination of locally adaptive approaches in contemporary machine learning systems. Section 3 details our proposed methodology, while Section 4 describes the implementation framework, experimental datasets—including our contributed robotics anomaly dataset—and comparative analysis of baseline methods against our proposed approach. Section 5 includes an ablation study that demonstrates the superiority of the interior-point method in terms of execution time. The paper concludes with final observations in Section 6.

\section{Prior Work}
Classifier fusion follows two main approaches: parallel, where multiple classifiers process inputs simultaneously, and sequential, where classifiers operate in a cascade. Both approaches can use either fixed rules or data-driven strategies for integration. While effective in multi-class scenarios, adapting these strategies to one-class classification (OCC) presents unique challenges, particularly in score calibration and decision boundary optimization due to the absence of counter-examples. This section analyzes parallel and sequential fusion methods, their application to OCC, and examines how locally adaptive techniques enhance these frameworks in modern machine learning systems.
\subsection{Parallel classifier fusion}
Parallel fusion architectures employ multiple independent classifiers operating concurrently, whose individual decisions are subsequently integrated to produce a final classification. This approach leverages the complementary strengths of diverse classifiers to effectively handle complex data distributions. The integration mechanisms can be categorized into fixed-rule and learning-based strategies, with the former detailed below.
\subsubsection{Fixed-rule Parallel Fusion}
Fixed-rule parallel fusion employs several strategies for combining classifier decisions \cite{kittler1998combining}, such as product, sum, max, min, median rules and majority voting. The product rule requires high confidence across all classifiers, while the sum rule assumes minimal deviation from prior probabilities, excelling in noisy scenarios. The max and min rules assign class membership based on highest probability output and highest minimum probability respectively. The median rule provides robustness against outliers, and majority voting determines class through predominant predictions. Each method offers unique advantages based on task requirements and data characteristics.
\subsubsection{Learning-based parallel fusion}
Learning-based parallel fusion integrates homogeneous or heterogeneous classifiers into ensemble frameworks, with stacking emerging as a prominent approach where meta-learners optimize the integration of base classifier predictions. Applications demonstrate its versatility: SVM meta-learners for object detection \cite{lisin2005combining}, neural networks for activity recognition \cite{rustam2020sensor}, deep networks for emotion recognition \cite{yin2017recognition}, and CNNs for medical imaging \cite{gour2022automated}. Variations include weighted averaging, successfully applied to face spoof detection, speech emotion recognition, and vehicle classification \cite{fatemifar2021particle, yalamanchili2022neural, hedeya2020super}.
\subsection{Sequential classifier fusion}
In contrast to parallel fusion architectures, sequential classifier systems implement a cascade methodology where base learners operate in succession, with each classifier's decision process being informed by and dependent upon the outputs of its predecessors in the processing pipeline.
\subsubsection{Fixed-rule sequential fusion}
Fixed-rule sequential fusion arranges classifiers in a cascade, using problem-specific heuristics. One approach evaluates match scores sequentially until reaching a confidence threshold \cite{poh2009benchmarking}. In biometric authentication \cite{marcialis2009personal}, a dual-threshold mechanism grants immediate authentication for scores above an upper threshold, rejects scores below a lower threshold, and defers intermediate cases to a secondary classifier. The effectiveness depends on appropriate threshold selection through heuristic methods.
\subsubsection{Learning-based sequential fusion}
Learning-based sequential fusion optimizes classifier integration using training data. Boosting \cite{schapire2003boosting} exemplifies this approach by reweighting misclassified samples to focus subsequent learners on difficult cases. Cascade generalization \cite{gama2000cascade} extends data at each level with probability distributions from base classifiers, though this creates higher-dimensional feature spaces that can complicate learning.
\subsection{Classifier Fusion for One-Class Learning}
One-class classification (OCC) focuses on learning from single-class samples to distinguish legitimate instances from anomalies. Building on ensemble methods' success in multi-class domains, researchers have developed various OCC-specific fusion strategies. These include weighted averaging for imbalanced scenarios \cite{FATEMIFAR20221}, conventional fusion rules for calibrated outputs \cite{tax2001combining}, and multiple SVDD classifiers for image analysis \cite{lai2002combining}. Advanced approaches encompass one-class SVM fusion with score normalization \cite{bergamini2009combining}, genetic algorithm optimization \cite{di2007combining}, stacking with model selection \cite{fatemifar2020stacking}, and specialized frameworks like KPCA integration \cite{zhang2014one} and one-class random forests \cite{desir2013one}. Applications span healthcare, biometrics, face spoofing detection, and medical image analysis.
\subsection{Locally Adaptive Learning}
Locally adaptive learning methods improve classification performance by adjusting decision boundaries based on local data characteristics. Early approaches like LLE and LBP \cite{roweis2000nonlinear, ojala2002multiresolution} introduced spatial adaptivity, while recent advances have expanded these capabilities. Modern approaches achieve superior results through manifold-aware kK-NN \cite{levada2024adaptive}, locally optimized k-NN \cite{pan2021new}, and Multiple Locally Linear Kernel Machines \cite{picard2024multiple}. For limited data scenarios, Few-Shot Node Classification \cite{xue2023few} and TALDS-Net \cite{qiao2024talds} demonstrate robust performance through local pattern optimization. These methods effectively handle complex real-world data while maintaining computational efficiency.

While existing approaches have made significant progress in classifier fusion and locally adaptive learning, several key challenges remain unaddressed. First, current fusion methods typically apply uniform weights across all data regions, failing to adapt to local data characteristics that may require different fusion strategies. Second, traditional optimization techniques for weight adjustment often suffer from computational inefficiency, limiting their applicability in real-time scenarios. Third, existing methods struggle to effectively balance between sparsity and uniformity in weight distribution, particularly when dealing with multiple diverse base classifiers. Our proposed method addresses these limitations through three key innovations: (1) the introduction of dynamic $\ell$p-norm constraints that adapt to local data patterns, (2) the implementation of an efficient interior-point optimization technique that significantly reduces computational overhead, and (3) the development of a locally adaptive learning framework that automatically adjusts fusion weights based on local data characteristics. These innovations enable our method to achieve superior performance while maintaining computational efficiency, as demonstrated in our experimental results.
\section{Methodology}
\label{met}

In real-world applications, it is observed that certain learners within an ensemble may exhibit superior discriminative abilities compared to their counterparts. Yet, prevalent approaches for fusing one-class classifiers often fall short in effectively managing this disparity. A notable deviation from this trend is the methodology introduced in \cite{FATEMIFAR20221}, which employs a genetic algorithm for the initial pruning of learners, followed by a phase dedicated to weight optimization. However, this method treats the pruning and weight fine-tuning as separate stages, potentially diluting the overall potency of the fusion approach. Furthermore, relying solely on a meta-heuristic for optimization lacks a guarantee for achieving the most favorable outcomes. In our previous work \cite{nourmohammadi2025lp}, we introduced a methodology that leverages the $\ell$p-norm constraint for one-class classifier fusion, enabling a sophisticated balance between sparsity and uniformity of the weights assigned to different classifiers in the ensemble.
\begin{eqnarray}
\nonumber \min_{\boldsymbol\omega}\sum_{i=1}^{n} \max(0,1-y_i\mathbf{s}_i^\top\boldsymbol{\omega}),&&\\
\text{s.t. } \lVert\boldsymbol{\omega}\rVert_p\leq 1,&&
\label{OF}
\end{eqnarray}
\noindent where $\lVert.\rVert_p$ denotes the $\ell_{p \ge 1}$-norm of a vector. For values of $p$ closer to $1$, the weight vector is expected to be relatively sparse while for larger values of $p$, especially for $p\rightarrow \infty$, the solution becomes more uniform \cite{nourmohammadi2025lp}.

This approach not only enhanced the discrimination power of the ensemble by selectively emphasizing more effective learners but also mitigated the risk of overfitting, thereby ensuring robust performance across diverse datasets. To address the computational challenges associated with this fusion method, we employed the Frank-Wolfe optimization technique, which was found to be more time-efficient than using conventional optimization packages like CVX. This choice significantly improved the computational efficiency of our learning process. Additionally, we incorporated hinge loss into our model, which outperformed the traditional least squares approach in detecting anomalies. This enhancement further underscored the practicality and effectiveness of our proposed method in real-world anomaly detection tasks.

However, a limitation of this method is its uniform treatment of samples, which does not adapt to local patterns within the data. Recognizing this limitation, our subsequent developments focus on enhancing model flexibility and sensitivity to variations across data sets:

\subsection{Locally Adaptive Learning with Dynamic \(\ell_p\)-norm Constraints}
This oversight can potentially limit the model's effectiveness in scenarios where local data characteristics significantly influence the overall classification performance. Our approach introduces a locally adaptive learning method, leveraging a dynamic adjustment of the \( \ell_p \)-norm constraint. This approach allows the model to adapt to local patterns in the data, leading to more refined decision boundaries and improved performance. Motivated by these insights, our investigation advocates for the adoption of a sparsity-inducing constraint aimed at harmonizing the influence spread across the ensemble’s members. This method, by determining sparsity directly from the data, improves the ability of certain learners to stand out compared to others and reduces the chance of overfitting or underfitting.

To incorporate local adaptiveness, we modify the traditional \( \ell_p \)-norm constraint by introducing a locality function \( L(\mathbf{x}_i, p) \), which dynamically adjusts the norm constraint \( p \) based on the local data characteristics \( \mathbf{x}_i \) (the feature vector of the \(i\)-th sample). Additionally, we allow for adaptive weight adjustments by introducing \( \boldsymbol{\omega}(\mathbf{x}_i) \), which represents the weight vector that varies for different subsets of data. The modified optimization problem is formulated as:

\begin{eqnarray}
\nonumber \min_{\boldsymbol{\omega}(\mathbf{x})} \sum_{i=1}^{n} && \max(0, 1 - y_i \mathbf{s}_i^\top \boldsymbol{\omega}(\mathbf{x}_i)) \\
&& \text{s.t. } \|\boldsymbol{\omega}(\mathbf{x}_i)\|_{L(\mathbf{x}_i, p)} \leq 1,
\label{OF}
\end{eqnarray}

\noindent where \( \mathbf{s}_i \) is the feature vector of the \(i\)-th training sample, \( \boldsymbol{\omega}(\mathbf{x}_i) \) is the weight vector associated with the local data characteristics \( \mathbf{x}_i \), and \( \|\cdot\|_{L(\mathbf{x}_i, p)} \) is the local \( \ell_p \)-norm adjusted by the locality function \( L(\mathbf{x}_i, p) \). The locality function adapts the norm constraint based on properties such as variability, allowing the model to dynamically respond to different regions of the data.

To manage the time execution more efficiently and address this computational bottleneck, we considered the Interior-Point optimization method, which is confirmed through comparative analysis to execute faster, providing a robust solution that enhances our method's scalability and applicability in larger, more complex datasets. This is due to its ability to handle constraints explicitly during the optimization process, which is critical when working with complex models that involve multiple base learners and a need for balance between sparsity and uniformity of the ensemble's weights. This characteristics allow the interior-point method to converge more rapidly, making it particularly well-suited for large-scale optimization problems where fast and reliable solutions are essential \cite{alizadeh1995interior}.

The goal of this approach is to optimize both the classifier's margin and the weight vector’s distribution, ensuring that the regularization and classifier weights are adaptive to local data characteristics. This is achieved through the introduction of the regularization parameter \( \mu \), which controls the trade-off between the classification margin and the norm of the weight vector.
\begin{eqnarray}
\min_{\boldsymbol{\omega}(\mathbf{x})} \left\{ \sum_{i=1}^{n} \max(0, 1 - y_i \mathbf{s}_i^\top \boldsymbol{\omega}(\mathbf{x}_i)) + \mu \|\boldsymbol{1-\omega}(\mathbf{x}_i)\|_{L(\mathbf{x}_i, p)} \right\}
\end{eqnarray}
    
The locality function \( L(\mathbf{x}_i, p) \) determines the local \( \ell_p \)-norm based on the sample’s characteristics \( \mathbf{x}_i \), allowing the model to become more flexible and locally adaptive. This formulation enhances local sensitivity while maintaining the overall structure of the ensemble classifier. When the locality function indicates more variability in the data, the model prefers a sparser set of weights (lower \( p \)-values), whereas in more uniform data regions, higher \( p \)-values are preferred, leading to a more balanced distribution of weights.

In summary, this locally adaptive learning mechanism dynamically adjusts both the regularization and the classifier weights based on local data insights, aiming to improve classification performance by finely tuning the model to the complexities of different data regions.

\subsection{Optimization with Locally Adaptive Interior-point Method}
To solve the convex optimization problem posed in Equation \ref{OF} with an \(\ell_p\)-norm constraint using the interior-point approach, we reformulate the problem into a form suitable for such methods. The objective function involves the hinge loss, which is piecewise linear and hence convex. The \( \ell_p \)-norm constraint is also convex for \( p \geq 1 \).

In the locally adaptive approach, the learning process adapts by dynamically adjusting the \( \ell_p \)-norm constraint based on the local data characteristics \( \mathbf{x}_i \) (the feature vector of the \( i \)-th sample). The weight vector \( \boldsymbol{\omega}(\mathbf{x}_i) \) varies across different subsets of data, allowing for greater flexibility in handling varying regions of the feature space. To enforce the locally adaptive \( \ell_p \)-norm constraint, we introduce a barrier function \( \phi(\boldsymbol{\omega}) \), which approaches infinity as \( \boldsymbol{\omega} \) nears the boundary of the feasible region defined by the \( \ell_p \)-norm constraint:

\begin{equation}
\phi(\boldsymbol{\omega}) = -\sum_{i=1}^{n} \ln(1 - \|\boldsymbol{\omega}(\mathbf{x}_i)\|_p^p)
\end{equation}

This barrier function ensures that the solution remains within the feasible region, i.e., \( \|\boldsymbol{\omega}(\mathbf{x}_i)\|_p < 1 \). The interior-point method solves the following locally adaptive optimization problem:

\begin{equation}
\min_{\boldsymbol{\omega}(\mathbf{x})} \left\{ \sum_{i=1}^{n} \max(0, 1 - y_i \mathbf{s}_i^\top \boldsymbol{\omega}(\mathbf{x}_i)) - \mu \sum_{i=1}^{n} \ln(1 - \|\boldsymbol{\omega}(\mathbf{x}_i)\|_p^p) \right\}
\end{equation}

\noindent where \( \mu \) is a positive parameter controlling the trade-off between the objective function and the barrier term. As \( \mu \) decreases, the solution of the barrier problem converges to the solution of the original problem.

The interior-point method offers improved efficiency compared to the Frank-Wolfe technique due to its ability to handle constraints directly during the optimization process. It transforms the constrained problem into an unconstrained one using a logarithmic barrier function, which simplifies the computational complexity and enables faster convergence. This method's explicit handling of constraints is particularly beneficial for complex models involving multiple learners, where balancing sparsity and uniformity in weight assignments is essential~\cite{alizadeh1995interior}.

The steps to solve the optimization problem using the interior-point method are (detailed in Algorithm \ref{alg:LocallyAdaptiveInteriorPointOptimization}):

\begin{enumerate}
    \item Choose an initial \( \mu > 0 \) and a starting point \( \boldsymbol{\omega}(\mathbf{x}_i) \) within the feasible region (\(\|\boldsymbol{\omega}(\mathbf{x}_i)\|_p < 1\)).
    \item Solve the barrier problem for the current value of \( \mu \) to obtain an approximate solution.
    \item Reduce \( \mu \) iteratively, solving the barrier problem from the previous solution.
    \item Apply the projection function to maintain adherence to the \( \ell_p \)-norm constraint after each iteration.
    \item Continue reducing \( \mu \) and optimizing until convergence, ensuring the weight vector \( \boldsymbol{\omega}(\mathbf{x}_i) \) remains within the feasible region.
\end{enumerate}

\subsubsection{Gradient Computation in the Locally Adaptive Space}
The gradient of the objective function with respect to \( \boldsymbol{\omega}(\mathbf{x}_i) \), including the barrier term, is given by:

\begin{equation}
\nabla_{\boldsymbol{\omega}(\mathbf{x}_i)} \phi(\boldsymbol{\omega}(\mathbf{x}_i)) = -\sum_{i=1}^{n} \frac{\nabla_{\boldsymbol{\omega}(\mathbf{x}_i)} \|\boldsymbol{\omega}(\mathbf{x}_i)\|_p}{1 - \|\boldsymbol{\omega}(\mathbf{x}_i)\|_p}
\label{gradf}
\end{equation}

The gradient \( \nabla_{\boldsymbol{\omega}(\mathbf{x}_i)} \|\boldsymbol{\omega}(\mathbf{x}_i)\|_p \) depends on the value of \( p \) and can be computationally complex for non-quadratic norms. This adaptive gradient computation ensures that the optimization process remains sensitive to local variations in the data.

A projection step ensures compliance with the \( \ell_p \)-norm constraint:

\begin{equation}
\min_{\boldsymbol{\omega}(\mathbf{x}_i)} \frac{1}{2} \|\boldsymbol{\omega}(\mathbf{x}_i) - \boldsymbol{\omega}^*\|_2^2 \quad \text{s.t.} \quad \|\boldsymbol{\omega}(\mathbf{x}_i)\|_p \leq 1
\end{equation}

By incorporating locally adaptive constraints into both optimization and projection steps, the interior-point method provides a robust and efficient framework for solving convex optimization problems with \( \ell_p \)-norm constraints. Its ability to adapt to local patterns in the data further enhances its performance in complex learning environments.

\begin{algorithm}
\caption{Locally Adaptive Optimization of Weights with Interior-Point Method for $\ell_p$-Norm Constraint}
\label{alg:LocallyAdaptiveInteriorPointOptimization}
\begin{algorithmic}[1]
\Require Score matrix $S \in \mathbb{R}^{n \times d}$; feature matrix $X \in \mathbb{R}^{n \times d}$; training labels $y \in \{-1, 1\}^n$; norm parameter $p \geq 1$; maximum number of epochs $max\_epochs$; learning rate $lr$.
\Ensure Optimized weights $W \in \mathbb{R}^{n \times d}$ for each locality.

\State Initialize $W \Leftarrow \{w_i \mid w_i = \frac{ones(d)}{norm(ones(d), p)}, \forall i\}$ \Comment{Initial weights normalized}
\State $\mu \Leftarrow 10$ \Comment{Barrier parameter for interior-point method}
\State $\beta \Leftarrow 0.5$ \Comment{Decay factor for $\mu$}
\State Define locality function $L(x_i, p)$ \Comment{Adjusts $p$ based on local data characteristics}

\For{$epoch = 1$ to $max\_epochs$}
    \ForAll{$i \in \{1, \dots, n\}$}
        \State $p_i \Leftarrow L(X[i], p)$ \Comment{Determine local norm parameter for sample $i$}
        \State $hinge\_loss_i \Leftarrow \max(0, 1 - y[i] \cdot (S[i, :] \cdot w_i))$ \Comment{Compute hinge loss}
        \State $active_i \Leftarrow (hinge\_loss_i > 0)$ \Comment{Check if loss is active}
        \State $gradient_i \Leftarrow -S[i, :]^T \cdot (y[i] \cdot active_i)$ \Comment{Compute gradient of hinge loss}

        \State \textbf{Compute Locally Adaptive Barrier Gradient:}
        \State $barrier\_term_i \Leftarrow -\ln(1 - norm(w_i, p_i)^{p_i})$
        \State $barrier\_gradient_i \Leftarrow -\mu \cdot \frac{p_i \cdot w_i^{(p_i - 1)} \cdot norm(w_i, p_i)^{p_i - 2}}{1 - norm(w_i, p_i)^{p_i}}$

        \State \textbf{Update Weights Locally:}
        \State $w_i \Leftarrow w_i - lr \cdot (gradient_i + barrier\_gradient_i)$

        \State \textbf{Project onto Locally Adaptive $\ell_{p_i}$-norm Ball:}
        \If{$norm(w_i, p_i) > 1$}
            \State $w_i \Leftarrow w_i \cdot \frac{1}{norm(w_i, p_i)}$ \Comment{Projection step}
        \EndIf
    \EndFor
    \State $\mu \Leftarrow \mu \cdot \beta$ \Comment{Decay barrier parameter for next iteration}
\EndFor

\State $W\_optimized \Leftarrow W$ \Comment{Finalize optimized weights}

\State \Return $W\_optimized$
\end{algorithmic}
\end{algorithm}

\subsection{Computational Cost}
The computational cost of Algorithm \ref{alg:LocallyAdaptiveInteriorPointOptimization} depends on the iteration count \(T\), training sample size \(n\), and score vector dimensionality \(d\). The introduction of locally adaptive learning impacts the computational complexity, primarily due to the gradient and barrier gradient computations tailored to each data point's local characteristics. The training phase complexity is \(O(T \cdot n \cdot d^2)\) due to the increased complexity of handling individual weight vectors and norm calculations for each sample. The main operations include:
\begin{itemize}
    \item \textbf{Gradient Computation:} \(O(n \cdot d)\) complexity for each data point's gradient.
    \item \textbf{Locally Adaptive Barrier Gradient Computation and Projection:} \(O(n \cdot d)\), similar to the gradient computation, since each weight vector \(w_i\) is adjusted and projected individually.
\end{itemize}
Despite the increased complexity, the efficiency of the testing phase remains significantly higher with a time complexity of \(O(m \cdot d)\), where \(m\) indicates the number of test samples. Although the training phase demands more computational resources due to the locally adaptive optimization, the testing phase benefits from a straightforward, non-iterative framework.

\subsection{Convergence}
The convergence of interior-point methods using a locally adaptive approach depends on the effective management of the \(\ell_p\)-norm constraint adjustments. For convex optimization problems, where the barrier function is self-concordant and locally adjusted, these methods can still exhibit quadratic convergence. However, due to the locality function, the number of iterations, \(T\), necessary for an \(\epsilon\)-approximation of the optimal solution \(\boldsymbol{\omega}^\star\) may vary. Specifically, with a geometric reduction of the barrier parameter \(\mu\) and the implementation of appropriate adaptive Newton steps, the method requires \(\mathcal{O}(\sqrt{n} \log(1/\epsilon))\) iterations to converge within an \(\epsilon\)-distance of \(\boldsymbol{\omega}^\star\), where \(n\) is the number of variables. The addition of locally adaptive constraints enriches the model's ability to finely tune to variations in data, potentially improving convergence characteristics in diverse and non-uniform datasets.

\subsection{Remarks}
Variations in the norm parameter \( p \) significantly affect the method's behavior. As \( p \) increases towards infinity, weights among classifiers uniformize, resembling the \texttt{sum} rule in classifier fusion. Conversely, as \( p \) approaches slightly above 1 (\( p \to 1^+\)), the method selectively prioritizes the classifier with the highest correct classifications, detailed in Eq. \ref{gradf}.

Setting \( p = 2 \) aligns the method with a soft-margin linear SVM, underlining its adaptability \cite{nourmohammadi2025lp}. Adjusting \( p \) within \([1, \infty)\) allows transitioning from focusing on the strongest classifier to equally weighting all, optimizing based on unique learner characteristics and data.

Assuming all classifiers surpass random performance, adjusting \( p \) to a more selective value can diminish the influence of under-performing classifiers, streamlining decision-making by reducing their weight.

\section{Experiments}
\label{exp}
This section presents our one-class classifier fusion method's evaluation across diverse datasets, comparing it with existing methods. We first detail implementation specifics (\S \ref{deta}), introduce the datasets (\S \ref{dats}), and discuss results (\S \ref{res})).
\subsection{Implementation Details} \label{deta} 
Our framework employs four base classifiers: Support Vector Data Description (SVDD), Kernelized one-class Gaussian Process (GP), Kernel PCA (KPCA), and One-class Gaussian Mixture Model (GMM). For SVDD, GP, and KPCA, we use RBF kernels with widths ${{0.25,0.5,1}}\times M$ ($M$: average squared Euclidean distance among training samples). KPCA's subspace dimension is selected from ${2,6,10,\dots,n}$, where $n$ represents the total training samples.

The preprocessing pipeline includes feature scaling, Z-score standardization of classifier scores, and two-sided min-max normalization (trimmed range technique, $\rho \in {1,\dots,10}$). The sparsity/uniformity parameter $p$ is chosen from \\$\{32/31,16/15,8/7,4/3,2,4,8,10,100\}$. For robotics datasets, features are extracted using pretrained CNNs (AlexNet, Xception, ResNet50, MobileNet).

\begin{table}[h!]
\scriptsize{
\setlength{\tabcolsep}{13pt} 
\renewcommand{\arraystretch}{1.5}
\caption{Descriptive statistics of datasets employed in experimental evaluation. *The Toyota HSR dataset is annotated at the video sequence level, with each complete sequence considered as one instance rather than frame-level labeling.}
\centering
\begin{tabular}{c l c c c }
\hline
&Data set & Samples & Normal & Features\\
\hline
D1&Banknotes & 1372 & 762 & 4\\
D2&Ionosphere & 351 & 225 & 34\\
D3&Vote & 435 & 267 & 16\\
D4&Glass & 214 & 76 & 9\\
D5&Iris & 150 & 50 & 4\\
D6&Breast Cancer Wisconsin & 699 & 458 & 9\\
D7&Wine & 178 & 71 & 13\\
D8 & Australia & 690 & 383 & 14\\
D9 & Haberman & 306 & 225 & 3\\
D10 & Hepatitis & 155 & 123 & 19\\
D11&Toyota HSR\cite{9636133}*&121 videos&61 videos&cf. \S \ref{dats}\\
D12&LiRAnomaly*&37076&31642&cf. \S \ref{lira_sec}\\
\hline
\end{tabular}
\label{datasets}}
\end{table}
\begin{figure}[h!]
\centering
\caption{Sequential frames depicting a Toyota HSR robot performing book manipulation tasks for anomaly detection evaluation.}
\includegraphics[width=.9\linewidth]{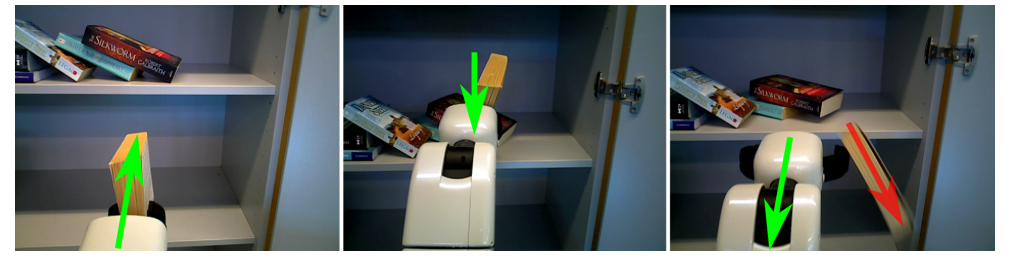}
\label{fig:anomaly-detection}
\end{figure}
\subsection{Dataset} 
\label{dats}
The evaluation uses 12 datasets (Table \ref{datasets}): ten UCI benchmark datasets and two specialized robotics datasets. The Toyota HSR dataset (Fig. \ref{fig:anomaly-detection}) captures book manipulation tasks with 48 nominal training trials, 6 validation sequences, and 67 test trials (60 anomalous), documenting various anomalies including book displacement, shelf disruption, sensor occlusion, and external perturbations.

\subsubsection{LiRAnomaly Dataset}
\label{lira_sec}
Our novel dataset, LiRAnomaly, comprises visual data collected from a Franka EMIKA robot during manipulation tasks, specifically focusing on pick-and-place operations between designated positions. The dataset encompasses 31,642 frames of normal operations and 5,434 frames containing anomalies, capturing both static and temporal anomalies that reflect critical operational failures and safety violations.
\begin{figure}[h!]
\begin{subfigure}{.3\textwidth}
\centering
\includegraphics[width=.9\linewidth]{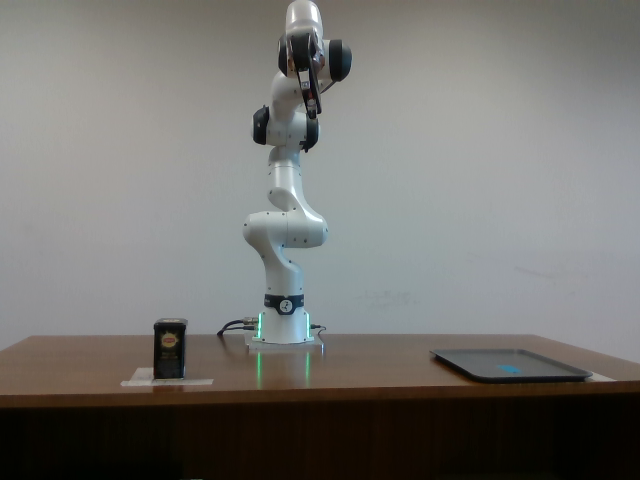}
\caption{Normal Operation}
\label{fig:sfig1}
\end{subfigure}%
\begin{subfigure}{.3\textwidth}
\centering
\includegraphics[width=.9\linewidth]{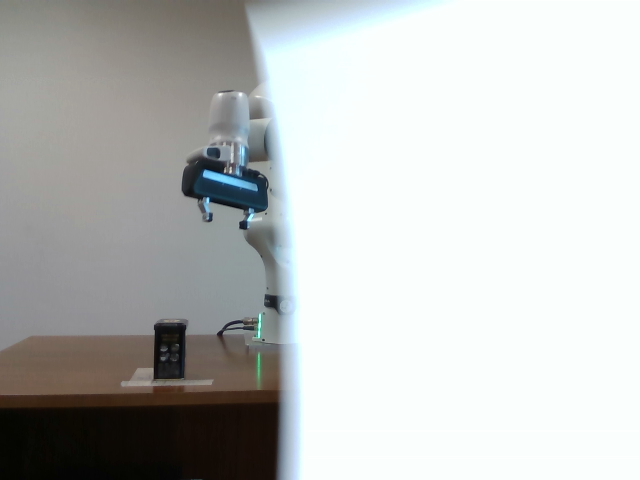}
\caption{Camera Occlusion}
\label{fig:sfig2}
\end{subfigure}
\begin{subfigure}{.3\textwidth}
\centering
\includegraphics[width=.9\linewidth]{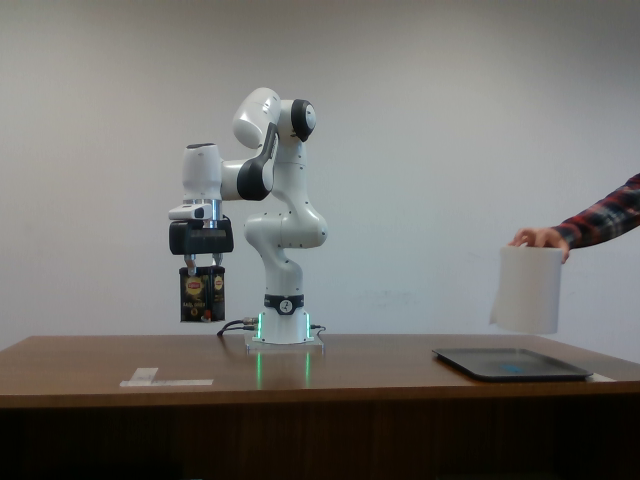}
\caption{Path Obstruction}
\label{fig:sfig2}
\end{subfigure}
\caption{Examples of normal operations and anomalous scenarios during box manipulation tasks.}
\label{fig:anomaly-detection-lira}
\end{figure}

The dataset documents four distinct categories of anomalies encountered during robotic operations. Type 1 anomalies represent visual sensor occlusion events, where the robot's vision system experiences temporary or partial blindness. Type 2 anomalies consist of grasp failures resulting from pose estimation errors, leading to unsuccessful object manipulation. Type 3 anomalies capture gripper malfunctions causing unintended object release during transport. Type 4 anomalies involve environmental obstacles appearing in the planned trajectory or target placement zone. These categories comprehensively cover the range of operational challenges encountered in robotic manipulation tasks.

% Our data partitioning strategy follows the protocols established in our previous work \cite{nourmohammadi2025lp}, a 70/20/10 split for normal samples in pure learning scenarios, with negative samples added to testing. Parameter optimization uses either RPAU measure \cite{fatemifar2022developing} or pseudo-negative generation (50\% of validation data)~\cite{Roth2006-di}. For non-pure scenarios, negative validation samples are split between training and validation. All experiments are repeated ten times for statistical validity, evaluated using AUC(ROC) and G-means metrics.
Our data partitioning strategy follows the protocols established in our previous work \cite{nourmohammadi2025lp}, using a 70/20/10 (train/validation/test) split for normal samples in pure learning scenarios, with negative samples added to testing. Parameter optimization employs two approaches: (1) The Relative Percentage Above Unity (RPAU) measure \cite{fatemifar2022developing}, which optimizes weighted average fusion using only normal validation data, and (2) a pseudo-negative generation technique \cite{Roth2006-di}, where 50\% of normal validation data is randomly selected and negatively replicated to simulate anomalous observations. For non-pure scenarios, negative validation samples are split between training and validation. All experiments are repeated ten times for statistical validity, evaluated using AUC(ROC) and G-means metrics.
\subsection{Results} \label{res} This section presents the results obtained from testing our one-class classifier fusion approach across various datasets and contrasts these findings with baseline and state-of-the-art methods.
 %%%%%%%%%%%%%%%%%%%%%%%%%%% RPAU %%%%%%%%%%%%%%%%%%%%%%
\begin{table}
\centering
\scriptsize{
\setlength{\tabcolsep}{1.5pt} 
\renewcommand{\arraystretch}{1.2}
\captionsetup{font=footnotesize}

\caption{Comparative analysis of the performances, represented by AUCs (mean$\pm$std\% in the upper multi-row) and G-means (mean$\pm$std\% in the lower multi-row), of the proposed `Pure' methods against various baselines and the sum rule across 10 iterations on UCI machine learning datasets, employing the RPAU measure for both Frank-Wolf and interior-point methods.}

\begin{threeparttable}[!h]
\resizebox{\columnwidth}{!}{
\begin{tabular}{ c c c c c c c c c c c c}
\hline
Data set &\multicolumn{1}{c}{Pure\tnote{1}}&\multicolumn{1}{c}{Pure\tnote{2}} & \multicolumn{1}{c}{GMM} & \multicolumn{1}{c}{SVDD} & \multicolumn{1}{c}{GP} & \multicolumn{1}{c}{KPCA} & \multicolumn{1}{c}{Sum rule} & \multicolumn{1}{c}{Single-best} & LOF \cite{10.1145/342009.335388} & IF \cite{4781136} & ALP \cite{lenz2021average} \\ \hline
         %& RPAU & pseudo-negative& & RPAU & pseudo-negative & RPAU & pseudo-negative & RPAU & pseudo-negative & RPAU & pseudo-negative & RPAU & pseudo-negative  & RPAU & pseudo-negative\\ \hline

\multirow{2}{*}{D1}&99.99$\pm$0.0&99.99$\pm$0.0& 83.90$\pm$4.5 & 92.90$\pm$1.0 &95.89$\pm$1.2& 80.85$\pm$2.2 &99.75$\pm$0.0 & 99.90$\pm$0.0 & 99.17$\pm$0.7 & 90.92$\pm$1.6 & 99.35$\pm$0.8 \\
                     &99.83$\pm$0.0&99.83$\pm$0.0& 90.51$\pm$0.2 & 92.80$\pm$0.2 & 93.89$\pm$0.7  & 74.01$\pm$5.8 & 97.00$\pm$0.0	& 99.34$\pm$0.0 & 92.83$\pm$1.1 & 80.99$\pm$2.8 & 90.72$\pm$2.2 \\
                     
\multirow{2}{*}{D2} &99.47$\pm$0.6& 99.49$\pm$0.6& 79.63$\pm$8.1 & 92.13$\pm$5.4  & 90.54$\pm$5.5 & 79.00$\pm$8.7 & 96.12$\pm$1.5& 97.49$\pm$1.5 & 95.72$\pm$2.8 & 89.94$\pm$3.2 & 95.99$\pm$2.2 \\
                     &98.32$\pm$1.5& 98.39$\pm$1.4 & 75.59$\pm$1.8 & 78.49$\pm$1.9 & 85.44$\pm$1.2 & 72.82$\pm$1.2 & 90.04$\pm$3.1& 92.28$\pm$2.2 & 86.77$\pm$4.9 & 80.56$\pm$3.7 & 78.66$\pm$6.2 \\
                     
\multirow{2}{*}{D3} &90.53$\pm$2.4& 92.20$\pm$2.6 & 71.70$\pm$2.3 & 78.04$\pm$2.5 & 74.16$\pm$4.3 & 66.27$\pm$1.7 & 80.72$\pm$3.0 & 90.34$\pm$5.4 & 93.74$\pm$1.7 & 96.59$\pm$1.2 & 94.05$\pm$2.0 \\
                     &83.43$\pm$3.0& 82.78$\pm$3.0 & 74.44$\pm$3.4 & 76.10$\pm$5.4 & 74.38$\pm$4.7 & 58.61$\pm$3.2 & 76.57$\pm$3.1 & 83.42$\pm$5.0 & 71.87$\pm$7.7 & 75.76$\pm$6.7 & 71.91$\pm$6.5 \\
\multirow{2}{*}{D4} &89.21$\pm$2.1& 92.75$\pm$2.0 & 79.34$\pm$6.5 & 83.15$\pm$3.5 & 81.88$\pm$5.3 &  65.77$\pm$9.1 & 90.03$\pm$2.5& 81.18$\pm$2.5 & 91.01$\pm$4.3 & 82.57$\pm$4.7 & 90.61$\pm$5.2 \\
                     &90.81$\pm$1.6& 92.86$\pm$1.7 & 79.80$\pm$8.0 & 81.42$\pm$7.0 & 86.08$\pm$6.5 & 80.08$\pm$6.0 & 85.97$\pm$4.5	& 70.29$\pm$3.4 & 77.58$\pm$9.5 & 65.76$\pm$7.7 & 69.29$\pm$4.5 \\
                     
\multirow{2}{*}{D5} &99.60$\pm$0.9&99.60$\pm$0.9& 96.20$\pm$2.2 & 96.50$\pm$2.3 & 97.00$\pm$2.8 & 94.00$\pm$1.6 & 98.45$\pm$1.0 & 99.20$\pm$0.0 & 92.00$\pm$7.7 & 93.56$\pm$6.3 & 92.20$\pm$7.5 \\
                     &98.99$\pm$1.0& 98.99$\pm$1.0 & 82.46$\pm$1.0 & 94.86$\pm$3.0 & 92.73$\pm$2.2 & 89.44$\pm$2.0 & 98.90$\pm$1.0 & 94.87$\pm$2.1 & 95.02$\pm$6.1 & 84.97$\pm$13.3 & 97.88$\pm$4.2 \\
                     
\multirow{2}{*}{D6} &98.98$\pm$1.0& 99.01$\pm$1.1& 82.83$\pm$3.0 & 94.83$\pm$3.0 & 95.51 $\pm$2.8 & 92.92$\pm$3.4 & 97.10$\pm$1.2 & 98.92$\pm$1.0 & 83.69$\pm$7.7 & 94.43$\pm$0.5 & 93.96$\pm$3.6 \\
                     &95.45$\pm$0.6& 95.45$\pm$0.6 & 85.63$\pm$3.8 & 93.44$\pm$4.0 & 94.65$\pm$3.3 & 89.44$\pm$4.0 & 95.38$\pm$1.2 &95.55$\pm$1.0 & 82.03$\pm$3.1 & 95.17$\pm$2.1 & 96.17$\pm$2.3 \\ 
                     
\multirow{2}{*}{D7} &91.64$\pm$3.0& 91.64$\pm$3.0& 73.85$\pm$6.1 & 85.84$\pm$6.0 & 83.01$\pm$7.7 & 75.20$\pm$5.0 & 86.79$\pm$2.0 & 92.86$\pm$1.1 & 95.53$\pm$3.2 & 95.18$\pm$3.2 & 95.81$\pm$3.0 \\
                     &89.92$\pm$1.6& 89.92$\pm$1.6 & 83.39$\pm$6.1 & 86.25$\pm$6.3 & 86.25$\pm$6.0 & 84.51$\pm$5.8 & 89.20$\pm$1.6 & 84.52$\pm$2.0& 73.48$\pm$5.8 & 86.45$\pm$6.2 & 72.93$\pm$4.7 \\
                     
\multirow{2}{*}{D8} & 79.80$\pm$2.8 & 79.85$\pm$2.6& 75.71$\pm$4.1 & 75.93$\pm$4.4 & 75.88$\pm$3.8 & 75.14$\pm$4.1 & 76.47$\pm$2.0 & 85.33$\pm$1.0 & 81.75$\pm$4.2 & 88.61$\pm$2.4 & 81.69$\pm$3.2 \\
                     &74.18$\pm$1.5& 75.29$\pm$1.5 & 66.07$\pm$2.0 & 69.39$\pm$2.1 & 70.67$\pm$2.0 & 69.86$\pm$2.2 & 72.65$\pm$1.8 & 78.50$\pm$0.8 & 78.15$\pm$2.8 & 81.24$\pm$3.1 & 79.10$\pm$2.9 \\
                     
\multirow{2}{*}{D9} & 66.81$\pm$4.8 & 66.59$\pm$5.0 & 57.72$\pm$1.0 & 57.84$\pm$0.9 & 58.40$\pm$1.6 & 53.06$\pm$0.8 & 61.20$\pm$5.0 & 58.28$\pm$3.1& 69.65$\pm$5.1 & 68.30$\pm$5.2 & 69.70$\pm$4.4 \\
                     &63.69$\pm$5.0& 64.04$\pm$5.0 & 56.40$\pm$1.1 & 58.29$\pm$1.2 & 58.38$\pm$1.4 & 54.77$\pm$1.1 & 61.72$\pm$3.0 & 57.74$\pm$4.0 & 63.77$\pm$4.3 & 60.90$\pm$5.4 & 60.00$\pm$5.0 \\
\multirow{2}{*}{D10} &77.60$\pm$2.8& 77.08$\pm$2.8 & 67.18$\pm$1.4 & 72.91$\pm$1.0 & 71.35$\pm$1.1 & 71.87$\pm$1.0 & 74.47$\pm$2.0 & 77.00$\pm$0.0 & 70.49$\pm$8.6 & 70.57$\pm$6.8 & 71.17$\pm$7.9 \\
                     &75.69$\pm$1.1& 74.80$\pm$1.2 & 68.46$\pm$2.4 & 70.71$\pm$2.2 & 57.73$\pm$2.1 & 66.14$\pm$2.0 & 71.80$\pm$3.4 & 71.42$\pm$1.0 & 66.22$\pm$4.8 & 66.02$\pm$7.5 & 62.95$\pm$9.3 \\
\hline
\end{tabular}}
     \begin{tablenotes}
     \item[1] Pure locally adaptive with frank-wolf.
     \item[2] Pure locally adaptive with interior-point.
   \end{tablenotes}
\end{threeparttable}%
\label{UCI_results_models_RPAU}}
\end{table}
 \subsubsection{Results on the UCI datasets}
Tables \ref{UCI_results_models_RPAU} and \ref{UCI_results_models_pseudo-negative} demonstrate the efficacy of our one-class classifier fusion methodology through comprehensive AUC and G-mean metrics (mean$\pm$std \%). Our evaluation framework encompasses ten trials on UCI datasets (Table \ref{datasets}), exploring various hyperparameter optimization scenarios using both RPAU measures and pseudo-negative samples. The methodology is benchmarked against foundational classifiers including GMM, SVDD, GP, KPCA, the sum-rule, and the highest-performing classifier determined through validation data.
%%%%%%%%%%%%%%% pseudo-negative %%%%%%%%%%%%%%%%%%%
\begin{table}
\centering
\scriptsize{
\setlength{\tabcolsep}{1.5pt} 
\renewcommand{\arraystretch}{1.2}
\captionsetup{font=footnotesize}
\caption{Comparative analysis of the performances, represented by AUCs (mean$\pm$std\% in the upper multi-row) and G-means (mean$\pm$std\% in the lower multi-row), of the proposed `Pure' methods against various baselines and the sum rule across 10 iterations on UCI machine learning datasets, employing the pseudo-negative samples for both Frank-Wolf and interior-point methods.}
\begin{threeparttable}[t]
\resizebox{\columnwidth}{!}{
\begin{tabular}{ c c c c c c c c c c c c}
\hline
Data set &\multicolumn{1}{c}{Pure\tnote{1}} & \multicolumn{1}{c}{Pure\tnote{2}} & \multicolumn{1}{c}{GMM} & \multicolumn{1}{c}{SVDD} & \multicolumn{1}{c}{GP} & \multicolumn{1}{c}{KPCA} & \multicolumn{1}{c}{Sum rule} & \multicolumn{1}{c}{Single-best} & LOF \cite{10.1145/342009.335388} & IF \cite{4781136} & ALP \cite{lenz2021average} \\ \hline
         %& RPAU & pseudo-negative& & RPAU & pseudo-negative & RPAU & pseudo-negative & RPAU & pseudo-negative & RPAU & pseudo-negative & RPAU & pseudo-negative  & RPAU & pseudo-negative\\ \hline

\multirow{2}{*}{D1}&99.99$\pm$0.0&99.99$\pm$0.0& 83.91$\pm$4.5 & 92.92$\pm$3.4 &  97.84$\pm$1.0 & 82.30$\pm$4.5 & 99.96$\pm$0.0 & 99.95$\pm$0.0 & 99.17$\pm$0.7 & 90.92$\pm$1.6 & 99.35$\pm$0.8 \\
                     & 99.83$\pm$0.0& 99.83$\pm$0.0 & 85.53$\pm$4.2 & 95.98$\pm$1.0 & 96.65$\pm$1.0 & 74.78$\pm$5.3 & 98.36$\pm$0.0 & 99.90$\pm$0.0 & 92.83$\pm$1.1 & 80.99$\pm$2.8 & 90.72$\pm$2.2 \\
                     
\multirow{2}{*}{D2} & 99.69$\pm$1.8 & 99.69$\pm$1.7& 79.65$\pm$7.9 & 94.37$\pm$4.8 & 89.39$\pm$6.0& 79.52$\pm$8.8 & 96.06$\pm$1.5 & 96.34$\pm$1.6 & 95.72$\pm$2.8 & 89.94$\pm$3.2 & 95.99$\pm$2.2 \\
                     &99.67$\pm$1.5& 99.67$\pm$1.5 & 76.63$\pm$2.0 & 89.08$\pm$1.4 & 86.78$\pm$1.2 & 80.01$\pm$2.0 & 90.77$\pm$3.3 & 90.72$\pm$2.1 & 86.77$\pm$4.9 & 80.56$\pm$3.7 & 78.66$\pm$6.2 \\
                     
\multirow{2}{*}{D3} &90.53$\pm$4.1& 90.53$\pm$4.0 & 81.97$\pm$3.0 & 81.88$\pm$2.8 & 83.80$\pm$3.0 & 62.65$\pm$2.4 & 84.78$\pm$3.3 & 94.03$\pm$5.0  & 93.74$\pm$1.7 & 96.59$\pm$1.2 & 94.05$\pm$2.0 \\
                     &84.07$\pm$4.0& 84.07$\pm$4.0 & 77.90$\pm$4.3 & 75.10$\pm$2.6 & 78.84$\pm$4.0 & 65.45$\pm$3.2 & 79.65$\pm$3.0 & 85.03$\pm$5.0 & 71.87$\pm$7.7 & 75.76$\pm$6.7 & 71.91$\pm$6.5 \\
                     
\multirow{2}{*}{D4} &92.75$\pm$2.2& 92.75$\pm$2.2 & 87.10$\pm$6.4 & 88.37$\pm$6.0 & 89.31$\pm$4.0 & 67.75$\pm$8.7 & 90.39$\pm$1.3 &  82.75$\pm$2.6 & 91.01$\pm$4.3 & 82.57$\pm$4.7 & 90.61$\pm$5.2 \\
                     &92.89$\pm$2.3& 92.89$\pm$2.3 & 79.62$\pm$8.0 & 85.97$\pm$7.3 & 82.75$\pm$6.0 & 69.15$\pm$7.0 & 89.38$\pm$1.6 & 73.03$\pm$2.1 & 77.58$\pm$9.5 & 65.76$\pm$7.7 & 69.29$\pm$4.5 \\

\multirow{2}{*}{D5} &99.60$\pm$0.9&99.60$\pm$1.0& 96.20$\pm$2.0 & 96.56$\pm$2.1 & 97.90$\pm$2.8 & 95.10$\pm$1.8 & 98.80$\pm$1.1 & 99.60$\pm$0.1 & 92.00$\pm$7.7 & 93.56$\pm$6.3 & 92.20$\pm$7.5 \\
                     &98.99$\pm$1.0& 98.99$\pm$1.0 & 82.46$\pm$1.0 & 88.54$\pm$1.2 & 95.99$\pm$2.3 & 90.55$\pm$ 2.0 & 98.99$\pm$1.0 & 95.92$\pm$2.0 & 95.02$\pm$6.1 & 84.97$\pm$13.3 & 97.88$\pm$4.2 \\
                     
\multirow{2}{*}{D6} &98.98$\pm$1.1& 99.03$\pm$1.1 & 82.00$\pm$3.0 & 95.67$\pm$3.2 & 95.12$\pm$3.0 & 92.57$\pm$3.0 & 97.01$\pm$1.0 & 98.89$\pm$1.0 & 83.69$\pm$7.7 & 94.43$\pm$0.5 & 93.96$\pm$3.6 \\
                     &95.45$\pm$0.6& 95.45$\pm$0.6 & 84.70$\pm$4.3 & 92.80$\pm$4.0 & 94.01$\pm$4.1 & 81.64$\pm$4.3 & 94.28$\pm$1.0 & 93.95$\pm$1.0 & 82.03$\pm$3.1 & 95.17$\pm$2.1 & 96.17$\pm$2.3 \\
                     
\multirow{2}{*}{D7} &91.64$\pm$2.2& 91.64$\pm$2.2& 84.94$\pm$6.6 & 85.98$\pm$4.1 & 86.79$\pm$7.3 & 87.60$\pm$4.3 & 88.14$\pm$1.2 & 94.44$\pm$1.1 & 95.53$\pm$3.2 & 95.18$\pm$3.2 & 95.81$\pm$3.0 \\
                     &89.99$\pm$1.0& 90.81$\pm$1.0 & 83.39$\pm$6.4 & 86.25$\pm$6.6 & 86.25$\pm$6.0 & 85.30$\pm$5.9 & 89.64$\pm$1.4 & 89.09$\pm$1.8 & 73.48$\pm$5.8 & 86.45$\pm$6.2 & 72.93$\pm$4.7 \\
                     
\multirow{2}{*}{D8} &79.80$\pm$2.0& 79.85$\pm$2.0 & 77.05$\pm$4.1 & 80.06$\pm$4.0 & 80.18$\pm$3.5 & 66.47$\pm$4.4 & 80.44$\pm$2.0 & 86.51$\pm$1.0 & 81.75$\pm$4.2 & 88.61$\pm$2.4 & 81.69$\pm$3.2 \\
                     &73.81$\pm$1.3& 73.91$\pm$1.4 & 67.69$\pm$2.0 & 69.59$\pm$1.8 & 65.87$\pm$2.0 & 70.01$\pm$2.3 & 73.40$\pm$2.0 & 79.18$\pm$1.0 & 78.15$\pm$2.8 & 81.24$\pm$3.1 & 79.10$\pm$2.9 \\
                     
\multirow{2}{*}{D9} &66.81$\pm$7.0& 66.59$\pm$7.1 & 57.90$\pm$1.0 & 58.02$\pm$1.1 & 60.25$\pm$1.5 & 54.00$\pm$1.0 & 62.70$\pm$5.0 & 58.71$\pm$5.0 & 69.65$\pm$5.1 & 68.30$\pm$5.2 & 69.70$\pm$4.4 \\
                     &63.69$\pm$5.2& 62.88$\pm$5.0 & 58.09$\pm$1.0 & 58.30$\pm$1.4 & 61.23$\pm$1.5 & 55.39$\pm$1.1 & 61.97$\pm$3.3 & 62.45$\pm$4.0 & 63.77$\pm$4.3 & 60.90$\pm$5.4 & 60.00$\pm$5.0 \\
                     
\multirow{2}{*}{D10} &77.60$\pm$1.4& 77.60$\pm$1.3& 67.18$\pm$1.4 & 70.72$\pm$0.8 & 74.79$\pm$1.0 & 72.91$\pm$1.0 & 75.52$\pm$2.3 & 77.09$\pm$2.0 & 70.49$\pm$8.6 & 70.57$\pm$6.8 & 71.17$\pm$7.9 \\
                     &75.69$\pm$1.2& 75.69$\pm$1.2& 68.46$\pm$2.2 & 70.71$\pm$2.0 & 67.70$\pm$2.0 & 66.14$\pm$2.0 & 73.37$\pm$3.0 &73.48$\pm$0.9 & 66.22$\pm$4.8 & 66.02$\pm$7.5 & 62.95$\pm$9.3 \\
\hline
\end{tabular}}
     \begin{tablenotes}
     \item[1] Pure locally adaptive with frank-wolf.
     \item[2] Pure locally adaptive with interior-point.
   \end{tablenotes}
\end{threeparttable}%
\label{UCI_results_models_pseudo-negative}}
\end{table}

The comparative analysis across datasets D1 to D10 reveals the robust performance of our 'Pure' methods against various baselines and the sum rule, utilizing both Frank-Wolf and interior-point optimization approaches. Notably, our methods demonstrate consistently superior AUC and G-mean scores compared to traditional approaches and modern baselines such as LOF, IF, and ALP. This performance advantage is particularly evident in the effective integration of our pure strategies compared to both the sum-rule and single-best classifier across diverse scenarios.
\begin{table}
\centering
\scriptsize{
\setlength{\tabcolsep}{0.1pt} 
\renewcommand{\arraystretch}{1.5}
\captionsetup{font=footnotesize}
\caption{Evaluation of the performances, denoted by AUCs (mean$\pm$std\% in the upper multi-row) and G-means (mean$\pm$std\% in the lower multi-row), of various versions of the proposed model (Pure' and Non-pure') compared to leading methods, conducted over 10 iterations on the UCI machine learning datasets presented in Table \ref{datasets}.}
\begin{threeparttable}[t]
\resizebox{\columnwidth}{!}{
\begin{tabular}{ c c c c c c c c c c c c c c c c c c c }
\hline
Data set  & \multicolumn{2}{c}{Pure\tnote{1}} & Non-pure &  \multicolumn{2}{c}{Pure\tnote{2}} & Non-pure & \multicolumn{3}{c}{Previous Work\cite{nourmohammadi2025lp}} &Fatemifar et al.\cite{fatemifar2022developing} & DEOD\tnote{3} \cite{wang2020dynamic}  & DEOD \tnote{4}\cite{wang2020dynamic}  & FNDF\cite{wang2022robust} & RAEOCSVMs\cite{xing2020robust}\\ %\cline{2-5}
 & RPAU & pseudo-negative & & RPAU & pseudo-negative& & RPAU & pseudo-negative & Non-pure &  \\ \hline

        \multirow{2}{*}{D1} &99.99$\pm$0.0&99.99$\pm$0.0&99.99$\pm$0.0&99.99$\pm$0.0&99.99$\pm$0.0&99.99$\pm$0.0& 99.97$\pm$0.0 & 99.97$\pm$0.0  & 99.98$\pm$0.0 & -& - & - & - & -\\
                               &99.83$\pm$0.0&99.83$\pm$0.0&99.91$\pm$0.0&99.83$\pm$0.0&99.83$\pm$0.0&99.91$\pm$0.0& 99.60$\pm$0.0 & 99.71$\pm$0.0 & 99.91$\pm$0.0 & 99.27 & - & - & - & 75.19\\
        \multirow{2}{*}{D2} &99.47$\pm$0.6&99.69$\pm$1.8&99.91$\pm$0.0&99.49$\pm$0.6&99.69$\pm$1.7&99.91$\pm$0.0& 97.42$\pm$1.6 & 97.45$\pm$1.7  & 98.39$\pm$1.1 & - & 84.53 & 85.93 & 94.90 & -\\
                               &98.32$\pm$1.5&99.67$\pm$1.5&99.73$\pm$0.8&98.39$\pm$1.4&99.67$\pm$1.5&99.73$\pm$1.1& 92.39$\pm$2.4 & 92.40$\pm$2.1 & 93.43$\pm$ 3.1 & 90.02 & - & - & 88.70 & -\\
        \multirow{2}{*}{D3}  &90.53$\pm$2.4&90.53$\pm$4.1&99.99$\pm$0.0&92.20$\pm$2.6&90.53$\pm$4.0&99.99$\pm$0.0& 85.86$\pm$3.6 & 86.07$\pm$3.0  & 99.99$\pm$0.0 & - & 90.76 & 92.06 & - & -\\
                              &83.43$\pm$3.0&84.07$\pm$4.0&98.55$\pm$2.1&82.78$\pm$3.0&84.07$\pm$4.0&98.55$\pm$2.0& 82.70$\pm$3.0 & 81.84$\pm$3.0 & 96.54$\pm$3.1 & 92.57 & - & - & - & -\\
        \multirow{2}{*}{D4}  &89.21$\pm$2.1&92.75$\pm$2.2&98.60$\pm$2.2&92.75$\pm$2.0&92.75$\pm$2.2&98.65$\pm$2.0& 91.12$\pm$1.0 & 91.66$\pm$1.3 & 95.21$\pm$3.1 & - & 83.00 & 84.34 & 62.1 & -\\
                               &90.81$\pm$1.6&92.89$\pm$2.3&94.01$\pm$1.3&92.86$\pm$1.7&92.89$\pm$2.3&94.64$\pm$1.4& 90.08$\pm$1.6 & 90.08$\pm$1.4 & 92/86$\pm$0.0 & - & - & - & 48.3 & 85.91\\
        \multirow{2}{*}{D5}  &99.60$\pm$0.9&99.60$\pm$0.9&99.98$\pm$0.0&99.60$\pm$0.9&99.60$\pm$1.0&99.98$\pm$0.0& 98.80$\pm$1.0 & 98.86$\pm$1.0 & 99.98$\pm$0.0 & - & 91.65 & 92.60 & 95.00 & -\\
                              &98.99$\pm$1.0&98.99$\pm$1.0&99.94$\pm$0.0&98.99$\pm$1.0&98.99$\pm$1.0&99.94$\pm$0.0& 98.99$\pm$1.0 & 91.00$\pm$1.0 & 99.90$\pm$0.0 & - & - & - & 90.70 & -\\
        \multirow{2}{*}{D6} &98.98$\pm$1.0&98.98$\pm$1.1&99.76$\pm$1.1 & 99.01$\pm$1.1&99.03$\pm$1.1&99.80$\pm$1.1& 97.27$\pm$0.8 & 98.96$\pm$1.0 & 99.39$\pm$0.4 & - & 90.77 & 90.60 & - & -\\
                               &95.45$\pm$0.6&95.45$\pm$0.6&99.15$\pm$0.8&95.45$\pm$0.6&95.45$\pm$0.6&99.22$\pm$0.8& 95.45$\pm$0.6 & 95.80$\pm$0.4 & 98.20$\pm$2.4 & 96.54 & - & - & - & -\\
        \multirow{2}{*}{D7}  &91.64$\pm$3.0&91.64$\pm$2.2&94.75$\pm$4.6&91.64$\pm$3.0&91.64$\pm$2.2&94.75$\pm$5.0& 88.94$\pm$2.7 & 91.01$\pm$1.4 & 94.17$\pm$4.3 & - & 81.04 & 78.90 & 90.8 & -\\
                               &89.92$\pm$1.6&89.99$\pm$1.0&92.44$\pm$6.0&89.92$\pm$1.6&90.81$\pm$1.0&92.44$\pm$4.4& 89.92$\pm$1.6 & 90.81$\pm$1.0 & 90.81$\pm$5.1 & - & - & - & 81.1 & -\\
        \multirow{2}{*}{D8} &79.80$\pm$2.8&79.80$\pm$2.0&81.30$\pm$0.4&79.85$\pm$2.6&79.85$\pm$2.0&81.30$\pm$0.3& 76.75$\pm$1.0 & 82.54$\pm$1.3  & 85.10$\pm$0.0 & - & - & - & - & -\\
                               &74.18$\pm$1.5&73.81$\pm$1.3&77.01$\pm$0.1&75.29$\pm$1.5&73.91$\pm$1.4&77.12$\pm$0.1& 73.81$\pm$0.1 & 74.92$\pm$1.2 & 79.42$\pm$0.0 & 71.45 & - & - & - & -\\
        \multirow{2}{*}{D9} &66.81$\pm$4.8&66.81$\pm$7.0&68.08$\pm$0.1&66.59$\pm$5.0&66.59$\pm$7.1&68.08$\pm$0.1& 67.61$\pm$5.0 & 67.61$\pm$5.2 & 69.61$\pm$0.0 & - & - & - & - & -\\
                               &63.69$\pm$5.0&63.69$\pm$5.2&64.50$\pm$0.4&64.04$\pm$5.0&62.88$\pm$5.0&64.83$\pm$0.4& 65.75$\pm$4.3 & 65.75$\pm$5.0 & 66.74$\pm$0.0 & 60.32 & - & - & - & -\\
        \multirow{2}{*}{D10} &77.60$\pm$2.8&77.60$\pm$1.4& 78.25$\pm$1.0&77.08$\pm$2.8&77.60$\pm$1.3&78.25$\pm$1.2& 76.56$\pm$1.1 & 78.52$\pm$1.2 & 78.64$\pm$0.0 & - & - & - & - & -\\
                             &75.69$\pm$1.1&75.69$\pm$1.2&85.20$\pm$0.1&74.80$\pm$1.2&75.69$\pm$1.2&85.06$\pm$0.0& 73.59$\pm$1.0 & 75.00$\pm$1.3 & 81.00$\pm$0.0 & 71.23 & - & - & - & 70.45\\
\hline
\end{tabular}}
     \begin{tablenotes}
          \item[1] Pure locally adaptive with frank-wolf.
     \item[2] Pure locally adaptive with interior-point.
     \item[3] Adaptive KNN homogeneous.
     \item[4] Adaptive KNN heterogeneous.
   \end{tablenotes}
    \end{threeparttable}%
\label{UCI_results_new}}
\end{table}
The Skillings–Mack statistical analysis test \cite{chatfield2009skillings}, is an extension of the Friedman test that is non-parametric. It is designed to handle situations where some methods may lack performance data. 
Further evaluation in Table \ref{UCI_results_new} according to the Skillings–Mack statistical analysis positions our pure methods against cutting-edge approaches, including our previous work \cite{nourmohammadi2025lp} and other approaches. 
%the  OCC ensemble optimization by Fatemifar {\em et al.} \cite{fatemifar2022developing}, the Dynamic Ensemble Outlier Detection (DEOD) system \cite{wang2020dynamic}, the Flexible Novelty Detection Framework (FNDF) \cite{wang2022robust}, and the Robust AdaBoost-based ensemble of one-class SVMs (RAEOCSVM) \cite{xing2020robust}.

The statistical analysis through Tables \ref{UCI_ranking_AUC_RPAU} to \ref{UCI_ranking_Gmeans} demonstrates the comprehensive effectiveness of our approach. The non-pure version, incorporating negative samples in training, achieves remarkable rankings of 2.20 and 1.85 in AUC and G-mean respectively. Similarly, the $\mathrm{Pure_{ps}}$ method, utilizing pseudo-negative samples, attains impressive rankings of 2.10 in AUC and 3.30 in G-mean, surpassing advanced methodologies including previous work \cite{nourmohammadi2025lp}.
\begin{table}[!h]
\renewcommand{\arraystretch}{1.0}
\centering
\scriptsize
\caption{Statistical comparison of different methods of Table \ref{UCI_results_models_RPAU} based on AUCs on different UCI datasets using the Skillings-Mac test (p-value=1.091e-7). Note: ``$\mathrm{._{r}}$" denotes using the RPAU for parameter tuning, infollows ``$\mathrm{Pure1_.}$" and ``$\mathrm{Pure2_.}$"  Frank-wolf and interior-point optimization techniques respectively.}
\begin{tabular}[h]{l c}
\toprule
    \textbf{Method} & \textbf{Mean ranking}  \\
\midrule               
    $\mathrm{GMM_{r}}$ & 10.00  \\
    $\mathrm{SVDD_{r}}$ & 7.10 \\
    $\mathrm{GP_{r}}$ & 7.40 \\
    $\mathrm{KPCA_{r}}$ & 9.90 \\
    $\mathrm{Sum-rule_{r}}$ & 5.10 \\
    $\mathrm{Single-best_{r}}$ & 4.40  \\
    LOF & 5.50\\
    IF & 5.80\\
    ALP & 4.70\\
    $\mathrm{Pure1_{r}(this\hspace{.1 cm} work)}$ & \textbf{3.35} \\
    $\mathrm{Pure2_{r}(this\hspace{.1 cm} work)}$ & \textbf{2.75} \\
\bottomrule
\label{UCI_ranking_AUC_RPAU}
\end{tabular}
\end{table}

\begin{table}[!h]
\renewcommand{\arraystretch}{1.0}
\centering
\scriptsize
\caption{Statistical comparison of different methods of Table \ref{UCI_results_models_RPAU} based on G-means on different UCI datasets using the Skillings-Mac test (p-value=9.309e-8). Note: ``$\mathrm{._{r}}$" denotes using the RPAU for parameter tuning, infollows ``$\mathrm{Pure1_.}$" and ``$\mathrm{Pure2_.}$"  Frank-wolf and interior-point optimization techniques respectively.}
\begin{tabular}[h]{l c}
\toprule
    \textbf{Method} & \textbf{Mean ranking}  \\
\midrule               
    $\mathrm{GMM_{r}}$ & 9.00  \\
    $\mathrm{SVDD_{r}}$ & 6.95 \\
    $\mathrm{GP_{r}}$ & 6.85 \\
    $\mathrm{KPCA_{r}}$ & 9.30 \\
    $\mathrm{Sum-rule_{r}}$ & 4.10 \\
    $\mathrm{Single-best_{r}}$ & 4.80  \\
    LOF & 6.80\\
    IF & 6.90\\
    ALP & 6.90\\
    $\mathrm{Pure1_{r}(this\hspace{.1 cm} work)}$ & \textbf{2.30} \\
    $\mathrm{Pure2_{r}(this\hspace{.1 cm} work)}$ & \textbf{2.10} \\
\bottomrule
\label{UCI_ranking_G-means_RPAU}
\end{tabular}
\end{table}
\clearpage
The Skillings-Mack test further validates our methods' superiority. The pure methods $\mathrm{Pure1_{r}}$ (Frank-Wolf) and $\mathrm{Pure2_{r}}$ (interior-point) achieve mean rankings of 3.35 and 2.75 respectively, significantly outperforming traditional approaches like GMM, SVDD, and KPCA. In G-means assessment, these methods lead with rankings of 2.30 and 2.10, demonstrating exceptional balance between sensitivity and specificity.
\begin{table}[!h]
\renewcommand{\arraystretch}{1.0}
\centering
\scriptsize
\caption{Statistical comparison of different methods of Table \ref{UCI_results_models_pseudo-negative} based on AUCs on different UCI datasets using the Skillings-Mac test (p-value=9.96e-7). Note: ``$\mathrm{._{ps}}$" denotes using the pseudo-negative for parameter tuning, infollows ``$\mathrm{Pure1_.}$" and ``$\mathrm{Pure2_.}$"  Frank-wolf and interior-point optimization techniques respectively.}
\begin{tabular}[!h]{l c}
\toprule
    \textbf{Method} & \textbf{Mean ranking}  \\
\midrule               
    $\mathrm{GMM_{ps}}$ & 9.70  \\
    $\mathrm{SVDD_{ps}}$ & 7.70    \\
    $\mathrm{GP_{ps}}$ & 6.80    \\
    $\mathrm{KPCA_{ps}}$ & 9.70   \\
    $\mathrm{Sum-rule_{ps}}$ & 4.90   \\
    $\mathrm{Single-best_{ps}}$ & 4.10   \\
    LOF & 5.70  \\
    IF & 6.00 \\
    ALP & 4.70\\
    $\mathrm{Pure1_{ps}(this \hspace{.1 cm} work)}$ & \textbf{3.40} \\
    $\mathrm{Pure2_{ps}(this \hspace{.1 cm} work)}$ & \textbf{3.30} \\
\bottomrule
\label{UCI_ranking_AUC_PS}
\end{tabular}
\end{table}
\begin{table}[!h]
\renewcommand{\arraystretch}{1.0}
\centering
\scriptsize
\caption{Statistical comparison of different methods of Table \ref{UCI_results_models_pseudo-negative} based on G-means on different UCI datasets using the Skillings-Mac test (p-value = 2.892e-8).  Note: ``$\mathrm{._{ps}}$" denotes using the pseudo-negative for parameter tuning, infollows ``$\mathrm{Pure1_.}$" and ``$\mathrm{Pure2_.}$"  Frank-wolf and interior-point optimization techniques respectively.}
\begin{tabular}[!h]{l c}
\toprule
    \textbf{Method} & \textbf{Mean ranking}  \\
\midrule               
    $\mathrm{GMM_{ps}}$ & 8.70  \\
    $\mathrm{SVDD_{ps}}$ & 6.95 \\
    $\mathrm{GP_{ps}}$ & 6.25 \\
    $\mathrm{KPCA_{ps}}$ & 9.60 \\
    $\mathrm{Sum-rule_{ps}}$ & 4.00 \\
    $\mathrm{Single-best_{ps}}$ & 4.00 \\
    LOF & 7.10 \\
    IF & 7.30 \\
    ALP & 7.40\\
    $\mathrm{Pure1_{ps}(this \hspace{.1 cm}work)}$ & \textbf{2.40} \\
    $\mathrm{Pure2_{ps}(this \hspace{.1 cm}work)}$ & \textbf{2.30} \\
\bottomrule
\label{UCI_ranking_G-means_PS}
\end{tabular}
\end{table}
\begin{table}[!h]
\renewcommand{\arraystretch}{1.0}
\centering
\scriptsize
\caption{Statistical comparison of the proposed methods with other approaches from Table \ref{UCI_results_new}, based on AUCs across different UCI datasets, using the Skillings-Mack test (p-value = 7.577e-9). Note: ``$\mathrm{._{r}}$" refers to using the RPAU and``$\mathrm{._{ps}}$'' refer to using pseudo-negative samples for parameter tuning. Infollows, ``$\mathrm{Pure1_.}$" and ``$\mathrm{Pure2_.}$" Frank-Wolfe and interior-point optimization techniques, respectively.}
\begin{tabular}[h]{l c}
\toprule
    \textbf{Method} & \textbf{Mean ranking}  \\
\midrule               
    DEOD\textsuperscript{3}\cite{wang2020dynamic} & 8.20\\
    DEOD\textsuperscript{4}\cite{wang2020dynamic} & 8.00 \\  
    FNDF \cite{wang2022robust} & 7.30\\
    Previous Work \cite{nourmohammadi2025lp}$\mathrm{Pure_{r}}$\cite{wang2020dynamic} & 8.70    \\
    Previous Work \cite{nourmohammadi2025lp}$\mathrm{Pure_{ps}}$\cite{wang2020dynamic} & 6.60 \\
   Previous Work \cite{nourmohammadi2025lp}$\mathrm{Non-pure}$\cite{wang2020dynamic} & 3.00 \\ 
    $\mathrm{Pure1_{r}(this\hspace{.1 cm}work)}$ & 6.40 \\
    $\mathrm{Pure1_{ps}(this\hspace{.1 cm}work)}$ & 5.75 \\
    $\mathrm{Pure2_{r}(this\hspace{.1 cm}work)}$ & 5.55 \\
    $\mathrm{Pure2_{ps}(this\hspace{.1 cm}work)}$ & 5.50 \\
    Non-Pure1 (this work)& \textbf{2.40} \\
    Non-Pure2 (this work)& \textbf{2.20} \\
\bottomrule
\label{UCI_ranking_AUC}
\end{tabular}
\end{table}

Our pseudo-negative sample-based approaches, $\mathrm{Pure1_{ps}}$ and $\mathrm{Pure2_{ps}}$, exhibit strong performance with AUC rankings of 3.40 and 3.30, and G-means rankings of 2.40 and 2.30 respectively. These results indicate that pseudo-negative incorporation can effectively approximate true negative sample performance, offering a viable solution when negative samples are unavailable.

The comprehensive evaluation establishes our mixed-sample methodology as a significant advancement in one-class classification. Through detailed statistical validation and consistent superior performance across diverse scenarios, our approach sets a new benchmark against established models and previous work \cite{nourmohammadi2025lp}, effectively leveraging locally adaptive optimization techniques to enhance classifier performance.
\clearpage
\begin{table}[!t]
\renewcommand{\arraystretch}{1.0}
    \centering
    \scriptsize
    \caption{Statistical comparison of the proposed methods with other approaches from Table \ref{UCI_results_new} based on G-means on different UCI datasets using the Skillings-Mack test (p-value = 6.256e-9 ). Note: ``$\mathrm{._{r}}$" refers to using the RPAU and``$\mathrm{._{ps}}$'' refer to using pseudo-negative samples for parameter tuning. Infollows, ``$\mathrm{Pure1_.}$" and ``$\mathrm{Pure2_.}$" Frank-Wolfe and interior-point optimization techniques, respectively.}
    \begin{tabular}[!h]{l c}
    \toprule
        \textbf{Method} & \textbf{Mean ranking}\\
    \midrule
        Fatemifar et al.\cite{fatemifar2022developing} & 7.50 \\
        FNDF \cite{wang2022robust} & 7.60\\  
        RAEOCSVMs\cite{xing2020robust} & 7.10 \\
        Previous Work \cite{nourmohammadi2025lp}$\mathrm{Pure_{r}}$\cite{wang2020dynamic} & 7.05    \\
        Previous Work \cite{nourmohammadi2025lp}$\mathrm{Pure_{ps}}$\cite{wang2020dynamic} & 6.70 \\
       Previous Work \cite{nourmohammadi2025lp}$\mathrm{Non-pure}$\cite{wang2020dynamic} & 3.15\\ 
        $\mathrm{Pure1_{r}(this\hspace{.1 cm}work)}$ & 6.60  \\
        $\mathrm{Pure1_{ps}(this\hspace{.1 cm}work)}$ & 5.90 \\
        $\mathrm{Pure2_{r}(this\hspace{.1 cm}work)}$ & 6.40 \\
        $\mathrm{Pure2_{ps}(this\hspace{.1 cm}work)}$ & 5.70 \\
        Non-Pure1 (this work)& \textbf{2.15} \\
        Non-Pure2 (this work)& \textbf{1.85} \\
    \bottomrule
    \label{UCI_ranking_Gmeans}
    \end{tabular}
\end{table}
\subsubsection{Results on the Toyota HSR and LiRAnomaly Dataset} 
The methodology employed in this paper \cite{9636133} involves a unique blend of a probabilistic U-Net architecture combined with analytical modeling of camera and body motions to detect deviations from expected behaviors during robotic task execution. This approach is characterized as a form of one-class classification, where the system learns from successful executions---capturing what constitutes normal behavior---and subsequently identifies deviations from these learned norms as anomalies. This method effectively combines data-driven machine learning techniques with traditional robotics modeling, allowing for effective anomaly detection even with smaller, task-specific datasets.

\begin{table}[!h]
\renewcommand{\arraystretch}{1.0}
    \centering
    \scriptsize
    \caption{
    Comparison of the proposed method variants against state-of-the-art [41] in terms of AUC-ROC (\%) and AUC-PR (\%). ".r" denotes the use of RPAU, and ".ps" denotes the use of pseudo-negative samples for parameter tuning. "Pure1." and "Pure2." represent the application of Frank-Wolfe and interior-point optimization techniques, respectively.Data sourced from \cite{9636133}}
    \begin{tabular}[t]{l c c }
    \toprule
        Method & \texttt{AUC-ROC} & \texttt{AUC-PR} \\
                           
    \midrule                   
        $\mathrm{Pure1_{r}(this\hspace{.1 cm}work)}$ &82.11  & 96.54 \\
        $\mathrm{Pure1_{ps}(this\hspace{.1 cm}work)}$& 82.15  & 96.55 \\
        $\mathrm{Pure2_{r}(this\hspace{.1 cm}work)}$& 82.11  & 96.55 \\
        $\mathrm{Pure2_{ps}(this\hspace{.1 cm}work)}$ &82.16  & 96.63 \\
        %$\mathrm{Non-pure1(this\hspace{.1 cm}work)}$ & 94.90  & 99.75  \\   
        %$\mathrm{Non-pure2(this\hspace{.1 cm}work)}$ & 94.90  & 99.75  \\  
        SOTA \cite{9636133} & 80.40 & 54.90\\
    \bottomrule
    \label{toyota}
    \end{tabular}
\end{table}

Table \ref{toyota} summarizes the performance comparison between our proposed method and the state-of-the-art (SOTA) [41]. The results demonstrate that our method, across all its variants (Pure1r, Pure1ps, Pure2r, Pure2ps), consistently outperforms the SOTA approach, achieving significantly higher scores in both AUC-ROC and AUC-PR metrics. Specifically, while the SOTA method achieves an AUC-ROC of 80.40\% and AUC-PR of 54.90\%, our method's variants achieve AUC-ROC scores of 82.11-82.16\% and AUC-PR scores of 96.54-96.63\%. This consistent improvement across all performance metrics demonstrates the robust detection capabilities of our proposed approach.

In anomaly detection tasks, particularly in non-pure novelty detection scenarios, having access to negative training samples does not guarantee their complete representation of test anomalies. This situation typically presents a straightforward binary classification challenge, where the objective is to discriminate between normal operations and various types of anomalies. This section uses the LiRAnomaly dataset, which includes four distinct types of operational anomalies, to evaluate the binary classification capabilities of our proposed method.

For this evaluation, our methodology is applied to train on normal operational conditions and a subset of known anomalies, and then it is tested against the full range of anomalies in the dataset. Table \ref{Lira} displays the average AUC (\%) for each type of anomaly, comparing our proposed method against existing techniques.

Our approach not only excels in accurately classifying different types of anomalies but also consistently achieves top performance across all categories, as evidenced by the highest rankings in each category from the LiRAnomaly dataset evaluation. As indicated by the results in Table \ref{Lira}, our method achieves AUC scores of 99.94\% for Type 1, 96.88\% for Type 2, 94.33\% for Type 3, and 92.98\% for Type 4 anomalies. The Non-pure version of our approach, which utilizes a mix of negative and positive training samples, notably outperforms all other tested methods, securing the top ranking with a statistical ranking of 1.00 across all anomaly types.

The Skillings-Mack statistical analysis, presented in Table \ref{Lira}, quantitatively validates the superiority of our method, with the lowest ranking scores indicating optimal performance. Statistical evaluation under controlled experimental conditions demonstrates that our approach significantly outperforms baseline methods (p-value = 0.00016) across all anomaly categories. The method exhibits particularly robust performance in challenging binary classification scenarios, achieving consistent detection rates across diverse anomaly types. The experimental results validate both the method's generalization capabilities and its statistical significance in handling complex anomaly detection tasks, with the Non-pure version achieving the highest possible ranking (1.00) across all evaluation metrics.

\begin{table}[!h]
\renewcommand{\arraystretch}{1.0}
    \centering
    \scriptsize
    \caption{Evaluation of different approaches on LiRAnomaly Dataset with four types of anomalies
    All results are presented in terms of AUC (\%) for low-quality video-level data. Statistical significance reported with a p-value of 0.00016. Note: ``$\mathrm{._{r}}$" refers to using the RPAU and``$\mathrm{._{ps}}$'' refer to using pseudo-negative samples for parameter tuning. Infollows, ``$\mathrm{Pure1_.}$" and ``$\mathrm{Pure2_.}$" Frank-Wolfe and interior-point optimization techniques, respectively.}
    \begin{tabular}[t]{l c c c c c }
    \toprule
        Method & \texttt{Type 1} & \texttt{Type 2} & \texttt{Type 3} & \texttt{Type 4} & Ranking \\
                           
    \midrule                   
        $\mathrm{KPCA_{r}}$ & 94.37 & 83.96 & 79.73 & 76.39 & 10.50\\
        $\mathrm{SVDD_{r}}$ & 98.98 & 89.34 & 87.03 & 88.93  & 5.50\\
        $\mathrm{GP_{r}}$& 98.83 & 91.96 & 80.75 & 83.33 & 7.00\\
        $\mathrm{GMM_{r}}$& 98.96 & 90.57 & 82.01 & 81.61 & 7.00 \\
        $\mathrm{KPCA_{ps}}$ & 94.37 & 83.96 & 79.73 & 76.39 & 10.50\\
        $\mathrm{SVDD_{ps}}$ & 98.98 & 89.34 & 87.03 & 88.93 & 5.50\\
        $\mathrm{GP_{ps}}$& 98.83 & 91.96 & 80.75 & 83.33 & 7.00\\
        $\mathrm{GMM_{ps}}$& 98.96 & 90.57 & 82.01 & 81.61 & 7.00\\
        $\mathrm{Pure1_{r}(this\hspace{.1 cm}work)}$& 99.81 & 96.73 & 93.20 & 92.63 &2.50    \\
        $\mathrm{Pure1_{ps}(this\hspace{.1 cm}work)}$& 99.81 & 96.73 & 93.20 & 92.63  &2.50    \\
        $\mathrm{Pure2_{r}(this\hspace{.1 cm}work)}$& 99.81 & 96.73 & 93.20 & 92.63  &2.50    \\
        $\mathrm{Pure2_{ps}(this\hspace{.1 cm}work)}$& 99.81 & 96.73 & 93.20 & 92.63  &2.50    \\
        $\mathrm{Non-pure(this\hspace{.1 cm}work)}$ & \textbf{99.94} & \textbf{96.88} & \textbf{94.33} & \textbf{92.98} & \textbf{1.00} \\        
    \bottomrule
    \label{Lira}
    \end{tabular}
\end{table}

\section{Ablation}

In our ablation study, we analyzed the efficiency of optimization methods within locally adaptive learning frameworks by comparing the execution times of the Interior Point and Frank-Wolf methods across different precision levels and specific values of $p$. Table~\ref{time} below outlines the improvements in execution times.

\begin{table}[!h]
\centering
\scriptsize
\renewcommand{\arraystretch}{1.5}
\caption{Comparison of execution times between Interior Point and Frank-Wolf methods, presented as multipliers indicating efficiency improvements.}
\label{time}
\begin{tabular}{lccc}
\hline
 & \textbf{precision=1e-2} & \textbf{precision=1e-3} & \textbf{precision=1e-4} \\
\hline
$p = \frac{32}{31}$ & 3.069$\times$ & 1.060$\times$ & 1.076$\times$ \\
$p = \frac{8}{7}$ & 2.257$\times$ & 1.442$\times$ & 1.142$\times$ \\
$p = 2$ & 3.360$\times$ & 6.510$\times$ & 2.052$\times$ \\
$p = 100$ & 3.716$\times$ & 2.344$\times$ & 19.003$\times$ \\
\hline
\label{time_execution}
\end{tabular}
\end{table}

The Interior Point method significantly outperforms the Frank-Wolf method in terms of computational speed, particularly as the precision requirements become more stringent. This can be seen most markedly in scenarios involving high $p$ values, where the execution time improvements reach up to $19.003\times$ for $p=100$ at a precision of $1e-4$. The superiority of the Interior Point method in these tests suggests that it can more quickly converge to an optimal solution, making it particularly suitable for applications in which time efficiency is critical. Such efficiency is pivotal for real-time systems or scenarios where rapid response is crucial.

These findings underscore the potential of the Interior Point method to enhance the computational performance of locally adaptive learning algorithms, particularly when handling large-scale datasets or complex optimization landscapes.

\section{Conclusion}
This research has presented a novel approach to one-class classifier fusion that leverages locally adaptive learning with dynamic $\ell$p-norm constraints. The proposed method demonstrates superior performance across diverse datasets, including both standard UCI benchmarks and specialized robotics applications like the Toyota HSR and LiRAnomaly datasets. The integration of the interior-point optimization method has shown significant computational efficiency improvements over the Frank-Wolfe technique, achieving up to 19 times faster execution in certain scenarios. The method's effectiveness is particularly notable in its ability to handle various types of anomalies, achieving high AUC scores (99.94$\%$, 96.88$\%$, 94.33$\%$, and 92.98$\%$) across different anomaly categories in the LiRAnomaly dataset. Statistical evaluations through the Skillings-Mack test confirm the method's superiority over existing approaches, with the non-pure version consistently achieving top rankings. The framework's ability to adapt to local data characteristics while maintaining computational efficiency makes it particularly valuable for real-world applications, especially in robotics and automation where rapid and accurate anomaly detection is crucial. These results establish the proposed method as a robust and practical solution for complex one-class classification challenges in both traditional and emerging application domains.

\section*{Acknowledgements}
This work was supported by TUBITAK under 2232 program with project number 121C148 (``LiRA").
\section*{Data statement}
The UCI datasets and Toyota HSR dataset used in this study are publicly available. The LiRAnomaly dataset will be made available upon reasonable request to the corresponding author.

\bibliography{ref}

\end{document}